\documentclass[preprint,12pt]{elsarticle}

\usepackage[utf8]{inputenc}
\usepackage{amsmath,amsfonts,amssymb,amsthm,bm,mathtools}
\usepackage{xcolor}
\usepackage{hyperref}
\usepackage{siunitx}
\usepackage{graphicx}
\usepackage{subcaption}
\usepackage{tabularray}
\usepackage{cleveref}

\DeclareMathOperator*{\argmin}{arg\,min}

\begin{document}
\begin{frontmatter}

\title{Statistical Learning Analysis of Physics-Informed Neural Networks}

\author[PNNL]{David Barajas-Solano\corref{label1}}
\ead{David.Barajas-Solano@pnnl.gov}
\cortext[label1]{Correspondence authors.}

\affiliation[PNNL]{organization={Pacific Northwest National Laboratory},%
city={Richland},
postcode={99352},
state={WA, USA}}

\begin{abstract}

  We study the training and performance of physics-informed learning for initial and boundary value problems (IBVP) with physics-informed neural networks (PINNs) from a statistical learning perspective.
  Specifically, we restrict ourselves to parameterizations with hard initial and boundary condition constraints and reformulate the problem of estimating PINN parameters as a statistical learning problem.
  From this perspective, the physics penalty on the IBVP residuals can be better understood not as a regularizing term bus as an infinite source of indirect data, and the learning process as fitting the PINN distribution of residuals \(p(y \mid x, t, w) q(x, t) \) to the true data-generating distribution \(\delta(0) q(x, t)\) by minimizing the Kullback-Leibler divergence between the true and PINN distributions.
  Furthermore, this analysis show that physics-informed learning with PINNs is a singular learning problem, and we employ singular learning theory tools, namely the so-called Local Learning Coefficient \cite{lau-llc-2025} to analyze the estimates of PINN parameters obtained via stochastic optimization for a heat equation IBVP.
  Finally, we discuss implications of this analysis on the quantification of predictive uncertainty of PINNs and the extrapolation capacity of PINNs.
  
\end{abstract}


\end{frontmatter}

\section{Introduction}

Physics-informed machine learning with physics-informed neural networks (PINNs) (see \cite{karniadakis-2021-physics} for a comprehensive overview) has been widely advanced and successfully employed in recent years to solve approximate the solution of initial and boundary value problems (IBVPs) in computational mathematics.
The central mechanism of physics-informed learning is model parameter identification by minimizing ``physics penalty'' terms penalizing the governing equation and initial and boundary condition residuals predicted by a given set of model parameters.
While various significant longstanding issues of the use PINNs for physics-informed learning such as the spectral bias have been addressed, it remains active area of research, and our understanding of various features of the problem such as the loss landscape, and the extrapolation capacity of PINN models, remains incomplete.

It has been widely recognized that traditional statistical tools are inadequate for analyzing deep learning.
Watanabe's ``singular learning theory'' (SLT) (see \cite{watanabe-algebraic-2009} for a review) recognizes that this is due to the singular character of learning with deep learning models and aims to address these limitations.
The development of efficient deep learning and Bayesian inference tools and frameworks \cite{homan-nuts-2014,welling-2011-sgld,heek-204-flax,cabezas-2024-blackjax} has enabled the application of SLT to analyzing practical applications of deep learning, leading to various insights on the features of the loss landscape, the generalization capacity of the model, and the training process \cite{lau-llc-2025,hoogland-2025-degeneracy,muhamed-2025-geometry}.

Given that physics-informed learning with PINNs is also singular due to its use of neural networks as the backbone architecture, we propose in this work to use SLT to analyze the training and performance of PINNs.
The physics penalty in physics-informed learning precludes an out-of-the-box application of SLT, so we first reformulate the PINN learning problem as a statistical learning problem by restricting our attention to PINN parameterizations with hard initial and boundary condition constraints.
Employing this formulation, it can be seen that the PINN learning problem is equivalent to the statistical learning problem of fitting the PINN distribution of governing equation residuals \(p(y \mid x, t, w) q(x, t) \) to the true data-generating distribution \(\delta(0) q(x, t)\) corresponding to zero residuals \( y(x, t) \equiv 0\) for all \( (x, t) \in \Omega \times (0, T] \), by minimizing the Kullback-Leibler divergence between the true and PINN distributions.
We argue this implies that the physics penalty in physics-informed learning can be better understood as an infinite source of indirect data (with the residuals interpreted as indirect observations given that they are a function of the PINN prediction) as opposed to as a regularizing term as it is commonly interpreted in the literature.
Furthermore, we employ the SLT's so-called Local Learning Coefficient (LLC) to study the features of the PINN loss for a heat equation IBVP.

This manuscript is structured as follows: In \cref{sec:slt} we reformulate and analyze the physics-informed learning problem as a singular statistical learning problem.
In \cref{sec:llc-numerical} we discuss the numerical approximation of the LLC and compute it for a heat equation IBVP.
Finally, further implications of this analysis are discussed in \cref{sec:discussion}.

\section{Statistical learning interpretation of physics-informed learning for IBVPs}
\label{sec:slt}

Consider the well-posed boundary value problem
\begin{align}
  \label{eq:ibvp-eq}
  \mathcal{L}(u(x, t), x, t) &= 0 && \forall x \in \Omega, && t \in (0, T],\\
  \label{eq:ibvp-bc}
  \mathcal{B}(u(x, t), x, t) &= 0 && \forall x \in \partial \Omega, && t \in [0, T],\\
  \label{eq:ibvp-ic}
  u(x, 0) &= u_0(x) && \forall x \in \Omega,
\end{align}
where the \(\Omega\) denotes the IBVP's spatial domain, \(\partial \Omega\) the domain's boundary, \(T\) the maximum simulation time, and \(u(x, t)\) the IBVP's solution.
In the PINN framework, we approximate the solution \(u(x, t)\) using an overparameterized deep learning-based model \(u(x, t, w)\), usually a neural network model \( \text{NN}_w(x, t) \), where \(w \in W\) denotes the model parameters and \(W \subset \mathbb{R}^d\) denotes the parameter space.
For simplicity, we assume that the model \(u(x, t, w)\) is constructed in such a way that the initial and boundary conditions are identically satisfied for any model parameter, that is,
\begin{gather*}
  \mathcal{B}(u(x, t, w), x, t) = 0 \quad \forall x \in \partial \Omega, \ t \in (0, T], \ w \in W,\\
  u(x, 0, w) = u_0(x) \quad \forall x \in \Omega, \ w \in W,
\end{gather*}
via hard-constraints parameterizations (e.g., \cite{li-2024-softhard,lu-2021-hard}).
The model parameters are then identified by minimizing the squared residuals of the governing equation evaluated at \(n\) ``residual evaluation'' points \( \{ x_i, t_i \}^n_{i = 1}\), that is, by minimizing the loss
\begin{equation}
  \label{eq:pinn-loss}
  L^{\text{PINN}}_n(w) \coloneqq \frac{1}{n} \sum^n_{i = 1} \left [ \mathcal{L}(u(x_i, t_i, w), x_i, t_i) \right ]^2,
\end{equation}

In the physics-informed-only limit, the PINN framework aims to approximate the solution without using direct observations of the solution, that is, without using labeled data pairs of the form \((x_i, t_i, u(x_i, t_i))\), which leads many authors to refer to it as an ``unsupervised'' learning framework.
An alternative view of the PINN learning problem is that it actually does use labeled data, but of the form \( (x_i, t_i, \mathcal{L}(u(x_i, t_i), x_i, t_i) \equiv 0 ) \), as the PINN formulation encourages the residuals of the parameterized model, \( \mathcal{L}(u(x, t, w), x, t) \), to be as close to zero as possible.
Crucially, ``residual'' data of this form is ``infinite'' in the sense that such pairs can be constructed for all \(x \in \Omega\), \(t \in (0, T]\).
In fact, the loss \labelcref{eq:pinn-loss} can be interpreted as a sample approximation of the volume-and-time-averaged loss \(L^{\text{PINN}} \coloneqq \int_{\Omega \times (0, T]} \left [ \mathcal{L}(u(x, t), x, t) \right ]^2 q(x, t) \, \mathrm{d} x \mathrm{d} t\) for a certain weighting function \(q(x, t)\).

We can then establish a parallel with statistical learning theory. Namely, we can introduce a density over the inputs, \(q(x, t)\), and the true data-generating distribution \(q(x, t, y) = \delta(0) q(x, t)\), where \(y\) denotes the equation residuals.
For a given error model \( p(y \mid \mathcal{L}(u(x, t, w), x, t)) \), the PINN model induces a conditional distribution of residuals \( p(y \mid x, w) \).
For example, for a Gaussian error model \( p(y \mid \mathcal{L}(u(x, t, w), x, t)) = \mathcal{N}(y \mid 0, \sigma^2) \), the conditional distribution of residuals is of the form
\begin{equation}
  \label{eq:pinn-distribution-residuals}
  p(y \mid x, t, w) \propto \exp \left \{ -\frac{1}{2 \sigma^2} \left [ \mathcal{L}(u(x, t, w), x, t) \right ]^2 \right \}.
\end{equation}
The PINN framework can then be understood as aiming to fit the model \(p(x, t, y \mid w) \coloneqq p(y \mid x, t, w) q(x, t) \) to the data-generating distribution by minimizing the sample negative log-likelihood \( L_n(w) \) of the training dataset \( \{ (x_i, t_i y_i \equiv 0) \}^n_{i = 1}\) drawn i.i.d. from \(q(x, t, y)\), with
\begin{equation}
  \label{eq:sample-nll}
  L_n(w) \coloneqq -\frac{1}{n} \sum^n_{i = 1} \log p(y_i \equiv 0 \mid x_i, t_i, w).
\end{equation}
Associated to this negative log-likelihood we also have the population negative log-likelihood \( L(w) \coloneqq -\mathbb{E}_{q(x, t, y)} \log p(y \mid x, t, w) \).
Note that for the Gaussian error model introduced above we have that \( L^{\text{PINN}}_n(w) = 2 \sigma^2 L_n(w) \) up to an additive constant, so that learning by minimizing the PINN loss \labelcref{eq:pinn-loss} and the negative log-likelihood \labelcref{eq:sample-nll} is equivalent.
Furthermore, also note that \( L^{\text{PINN}}(w) = 2 \sigma^2 L(w) \) up to an additive constant if the volume-and-time averaging function is taken to be the same as the data-generating distribution of the inputs, \(q(x, t)\).
Therefore, it follows that PINN training is equivalent to minimizing the Kullback-Leibler divergence between the true and model distributions of the residuals.

This equivalent reformulation of the PINN learning problem allows us to employ the tools of statistical learning theory to analyze the training of PINNs.
Specifically, we observe that, because of the neural network-based forward model \( \text{NN}_w(x) \) is a \emph{singular} model, the PINN model \( p(y \mid x, t, w) \) is also singular, that is, the parameter-to-distribution map \( w \mapsto p(x, t, y \mid w) \) is not one-to-one, and the Fisher information matrix \(I(w) \coloneqq \mathbb{E}_{p(x, t, y \mid w)} \nabla \nabla_w \, \log p(x, t, y \mid w) \) is not everywhere positive definite in \(W\).
As a consequence, the PINN loss \labelcref{eq:pinn-loss} is not characterized by a discrete, measure-zero, possibly infinite set of local minima that can be locally approximated quadratically.
Instead, the PINN loss landscape, just as is generally the case for deep learning loss landscapes, is characterized by ``flat'' minima, crucially \emph{even in the limit of the number of residual evaluation points \(n \to \infty\)}.
This feature allows us to better understand the role of the residual data in physics-informed learning.

Namely, it has been widely observed that the solutions to the PINN problem \(\argmin_w L^{\text{PINN}}_n(w) \) for a given loss tolerance \(\epsilon\) are not unique even for large number of residual evaluation points, and regardless of the scheme used to generate such points (uniformly distributed, low-discrepancy sequences such as latin hypercube sampling, random sampling, etc.), while on the other hand it has also been observed that increasing \(n\) seems to ``regularize'' or ``smoothen'' the PINN loss landscape (e.g., \cite{tartakovsky-pinn-2024}).
Although one may interpret this observed smoothing as similar to the regularizing effect of explicit regularization (e.g., \(\ell_2\) regularization or the use of a proper prior on the parameters), in fact increasing \(n\) doesn't sharpen the loss landscape into locally quadratic local minima as explicit regularization does.
Therefore, residual data (and correspondingly, the physics penalty) in physics-informed learning should not be understood as a regularizer but simply as a source of indirect data.

As estimates of the solution to the PINN problem are not unique, it is necessary to characterize them through other means besides their loss value.
In this work we employ the SLT-based LLC metric, which aims to characterize the ``flatness'' of the loss landscape of singular models.
For completeness, we reproduce the definition of the LLC (introduced in \cite{lau-llc-2025}) in the sequel.
Let \(w^\star\) denote a local minima and \(B(w^\star)\) a sufficiently small neighborhood of \( w^\star\) such that \( L(\omega) \geq L(\omega^\star) \); furthermore, let \( B(w^\star, \epsilon) \coloneqq \{ w \in B(w^\star) \mid L(w) - L(w^\star) < \epsilon\}\) denote the subset of the neighborhood of \(w^\star\) with loss within a certain tolerance \(\epsilon\), and \(V(\epsilon)\) denote the volume of such subset, that is,
\begin{equation*}
  V(\epsilon) \coloneqq \operatorname{vol}( B(w^\star, \epsilon) ) = \int_{B(w^\star, \epsilon)} \mathrm{d} w.
\end{equation*}
Then, the LLC is defined as the rational number \(\lambda(w^\star)\) such that the volume scales as \(V(\epsilon) \propto \exp \{ \lambda(w^\star) \} \).
The LLC can be interpreted as a measure of flatness because it corresponds to the rate at which the volume of solutions within the tolerance \(\epsilon\) decreases as \(\epsilon \to 0\), with smaller LLC values correspond to slower loss of volume.

As its name implies, the LLC is the local analog to the SLT (global) learning coefficient \(\lambda\), which figures in the asymptotic expansion of the free energy of a singular statistical model.
Specifically \(\lambda(w^\star) \equiv \lambda\) if \(w^\star\) is a global minimum and \(B(w^\star)\) is taken to be the entire parameter space \(W\).
The free energy \(F_n\) is defined as
\begin{equation*}
  F_n \coloneqq -\log \int \exp \{ -n L_n(w) \} \varphi(w) \, \mathrm{d} w,
\end{equation*}
where \(\varphi(w)\) is a prior on the model parameters.
Then, according to the SLT, the free energy for a singular model has the asymptotic expansion \cite{watanabe-algebraic-2009}
\begin{equation*}
  F_n = nL_n(w_0) + \lambda \log n + O_P(\log \log n), \quad n \to \infty,
\end{equation*}
where \(w_0\) is the parameter vector that minimizes the Kullback-Leibler divergence between the true data-generating distribution and the estimated distribution \( p(y \mid x, t, w) q(x, t) \), whereas for a regular model it has the expansion
\begin{equation*}
  F_n = n L_n(\hat{w}) + \frac{d}{2} \log n + O_P(1), \quad n \to \infty,
\end{equation*}
where \(\hat{w}\) is the maximum likelihood estimator of the model parameters, and \(d\) the number of model parameters.
The learning coefficient can then be interpreted as measuring model complexity, as it is the singular model analog to a regular model's model complexity \(d / 2\).
Therefore, by computing the LLC around different estimates to the PINN solution we can gain some insight into the PINN loss landscape.

\section{Numerical estimation of the local learning coefficient}
\label{sec:llc-numerical}

\subsection{MCMC-based local learning coefficient estimator}

While there may exist other approaches to estimating the LLC (namely, via Imai's estimator \cite{imai-estimating-2019} or Watanabe's estimator \cite{watanabe-wbic-2013} of the real log-canonical threshold), in this work we choose to employ the estimation framework introduced in \cite{lau-llc-2025}, which we briefly describe as follows.
Let \(B_\gamma(w^\star)\) denote a small ball of radius \(\gamma\) centered around \(w^\star\). We introduce the local to \(w^\star\) analog to the free energy and its asymptotic expansion:
\begin{equation}
  \label{eq:local-free-energy}
  \begin{split}
    F_n(B_\gamma(w^\star)) &= -\log \int_{B_\gamma(w^\star)} \exp \{ -n L_n(w) \} \varphi(w) \, \mathrm{d} w \\
    &= n L_n(w^\star) + \lambda(w^\star) \log n + O_P(\log \log n).
  \end{split}
\end{equation}
We then substitute the hard restriction in the integral above to \(B_\gamma(w^\star)\) with a softer Gaussian weighting around \(w^\star\) with covariance \(\gamma^{-1} I_d \), where \(\gamma > 0 \) is a scale parameter.
Let \(\mathbb{E}_{w \mid w^\star, \beta, \gamma} [ \cdot ]\) denote the expectation with respect to the tempered distribution
\begin{equation*}
  p(w \mid w^\star, \beta, \gamma) \propto \exp \left \{ - n \beta L_n(w) - \frac{\gamma}{2} \| w - w^\star \|^2_2 \right \}
\end{equation*}
with inverse temperature \(\beta \equiv 1 / \log n\); then, \(\mathbb{E}_{w \mid w^\star, \beta, \gamma} [ n L_n(w) ]\) is a good estimator of the partition function of this tempered distribution, which is, depending on the choice \(\gamma\), an approximation to the local free energy.
Substituting this approximation into \labelcref{eq:local-free-energy}, we obtain the estimator
\begin{equation}
  \label{eq:llc-estimator}
  \hat{\lambda}_\gamma(w^\star) \coloneqq n \beta \left [ \mathbb{E}_{w \mid w^\star, \beta, \gamma} [ L_n(w) ] - L_n(w^\star) \right ].
\end{equation}

In this work we compute this estimate by computing samples \( w \mid w^\star, \beta, \gamma \) via Markov chain Monte Carlo (MCMC) simulation.
Specifically, we employ the No-U Turns Sampler (NUTS) algorithm \cite{homan-nuts-2014}.
Furthermore, to evaluate \(L_n(w)\) we choose a fixed set of \(n\) residual evaluation points chosen randomly from \(q(x, t)\), with \(n\) large enough that we can employ the asymptotic expansion \labelcref{eq:local-free-energy} and that \cref{eq:llc-estimator} is valid.
Note that \(\hat{\lambda}_\gamma(w^\star) \) is strictly positive as \(w^\star\) is a local minima of \(L_n(w)\), so that \(\mathbb{E}_{w \mid w^\star, \beta, \gamma} [ L_n(w) ] - L_n(w^\star) \geq 0\).
In fact, negative estimates of the LLC imply that the MCMC chain was able to visit a significant volume of parameter space for which \( L_n(w) < L_n(w^\star) \), which means that \(w^\star\) is not an actual minima of \(L_n(w)\).
Given that minimization of the PINN loss via stochastic optimization algorithms is not guaranteed to produce exact minima, in practice it may be possible to obtain negative estimates of the LLC.

\subsection{Numerical experiments}

To analyze the PINN loss landscape for a concrete case, we compute and analyze the LLC for the following physics-informed learning problem:
We consider the heat equation IBVP
\begin{align*}
  \partial_t u - \partial_{xx} u &= f(x, t) && \forall x \in (0, 2), \ t \in (0, 2],\\
  u(0, t) = u(2, t) &= 0 && \forall t \in (0, 2],\\
  u(x, 0) &= \sin(\pi x) && \forall x \in (0, 2),
\end{align*}
where \(f(x, t)\) is chosen so that the solution to the IBVP is
\begin{equation*}
  u(x, t) = \exp^{-t} \sin(\pi x).
\end{equation*}
To exactly enforce initial and boundary conditions, we employ the hard-constraints parameterization of the PINN model
\begin{equation}
  \label{eq:hard-constraints}
  u(x, t, w) = \sin(\pi x) + t x (x - 1) \text{NN}_w(x, t).
\end{equation}
The neural network model \(\text{NN}_w\) is taken to be a multi-layer perceptron (MLP) with 2 hidden layers of \(50\) units each.
This model has a total number of parameters \(d = 20,601.\)
This relatively small model ensures that the wall-clock time of computing multiple MCMC chains is reasonable.
The model \cref{eq:hard-constraints} is implemented using \textsc{Flax NNX} library \cite{heek-204-flax}, and MCMC sampling is performed using the \textsc{BlackJAX} library \cite{cabezas-2024-blackjax}.

\begin{figure}[th]
  \centering
  \begin{subfigure}[b]{0.48\textwidth}
    \includegraphics[width=\textwidth, trim={0 0.4cm 0 0}, clip]{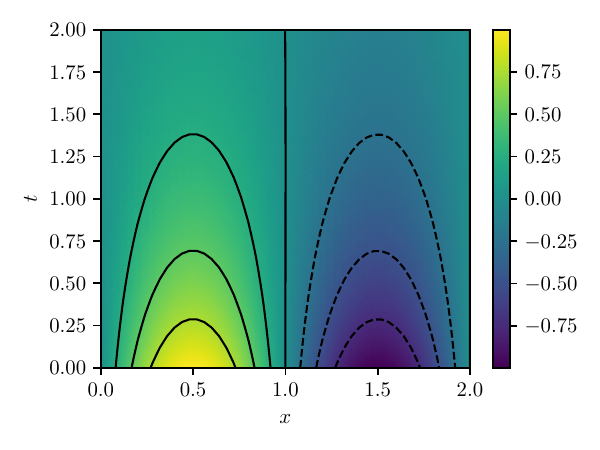}
    \caption{PINN solution \(u(x, t, w^\star)\)}
    \label{fig:pred}
  \end{subfigure}
  \centering
  \begin{subfigure}[b]{0.48\textwidth}
    \includegraphics[width=\textwidth, trim={0 0.4cm 0 0}, clip]{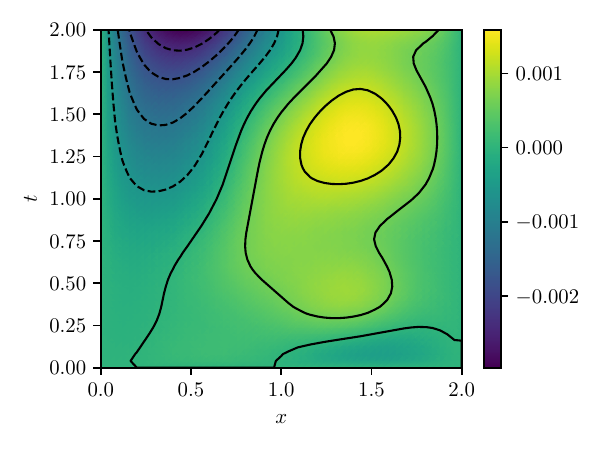}
    \caption{Pointwise errors \(u(x, t) - u(x, t, w^\star)\)}
    \label{fig:error}
  \end{subfigure}
  \caption{PINN solution and estimation error computed with batch size of \(32\) and Adam learning rate of \(1 \times 10^{-4}.\)}
  \label{fig:sol}
\end{figure}

The parameters of the model are estimated by minimizing the PINN loss \cref{eq:pinn-loss} using the Adam stochastic optimization algorithm for \(50,000 \) iterations.
At each iteration, a new batch of residual evaluation points is drawn from \(q(x, t)\) and used to evaluate the PINN loss and its gradient.
Note that these batches of residual evaluation points are distinct from the set used to evaluate \(L_n(w)\) in \cref{eq:llc-estimator}, which is fixed for a given experiment.
\Cref{fig:sol} shows the PINN solution \( u(x, t, w^\star) \) for \(w^\star\) estimated using a batch size of \(32\) and learning rate of \(1 \times 10^{-4}\).
This solution has an associated PINN loss of \(1.854 \times 10^{-5}\), which, although small, shows that the true data-generating distribution \(\delta (0) q(x, t)\) is not realizable within the model class \labelcref{eq:hard-constraints} with the MLP architecture indicated above due to the finite capacity of the class.

LLC estimates are computed for batch sizes of \(8\), \(16\), and \(32\), and for Adam learning rates \(\eta\) of \(1 \times 10^{-3}\) and \(1 \times 10^{-4}.\)
Furthermore, for all experiments we compute the LLC estimate using a fixed set of \(256\) residual evaluation points drawn randomly from \(q(x, t).\)
To compute the LLC, we employ the Gaussian likelihood of residuals \eqref{eq:pinn-distribution-residuals}.
As such, the LLC estimate depends not only on the choice of \(\gamma\) but also on the standard deviation \(\sigma\).
For each experiment we generate 2 MCMC chains and each chain is initialized with an warmup adaptation interval.

To inform the choice of hyperparameters of the LLC estimator we make the following observations:
First, both hyperparameters control the volume of the local neighborhood employed to estimate the LLC, in the case of \(\gamma\) by expanding or contracting the prior centered around \(w^\star\), and in the case of \(\sigma\) by increasing or decreasing the range of values \(L_n(w)\) can take within the local neighborhood.
Second, given that the true data-generating distribution is not realizable, the PINN loss will always be non-zero.
Therefore, for the true zero residuals to be resolvable with non-trivial probability by the PINN model we must choose \(\sigma\) larger than the average optimal PINN residuals (the squared root of the optimal PINN loss.)

Given that \(\gamma = 1\) controls the volume of the local neighborhood, we take the value of \(\gamma = 1\)
For the architecture described above, we find that the optimal PINN loss is \(O(10^{-5})\), so that \(\sigma\) must be chosen to be \(3 \times 10^{-3}\) or larger.
Given that \(\sigma\) also controls the volume of the local neighborhood, we take the conservative value of \(\sigma = 1\).
It is not entirely obvious whether this conservative choice overestimates or underestimates the LLC.
Numerical experiments not presented in this work showed that the small choice \(\sigma = 1 \times 10^{-2}\) led often to negative LLC estimates and therefore was rejected.
LLC values estimated for \( \sigma = 1 \times 10^{-1} \) where consistently slightly larger than those for \(\sigma = 1\), which indicates that the model class of PINN models that are more physics-respecting has slightly more complexity than a less physics-respecting class.
Nevertheless, given that the difference in LLC estimates is small we proceeded with the conservative choice.

\begin{figure}
  \centering
  \begin{subfigure}[b]{\textwidth}
    \includegraphics[width=\textwidth, trim={0 0.4cm 0 0}, clip]{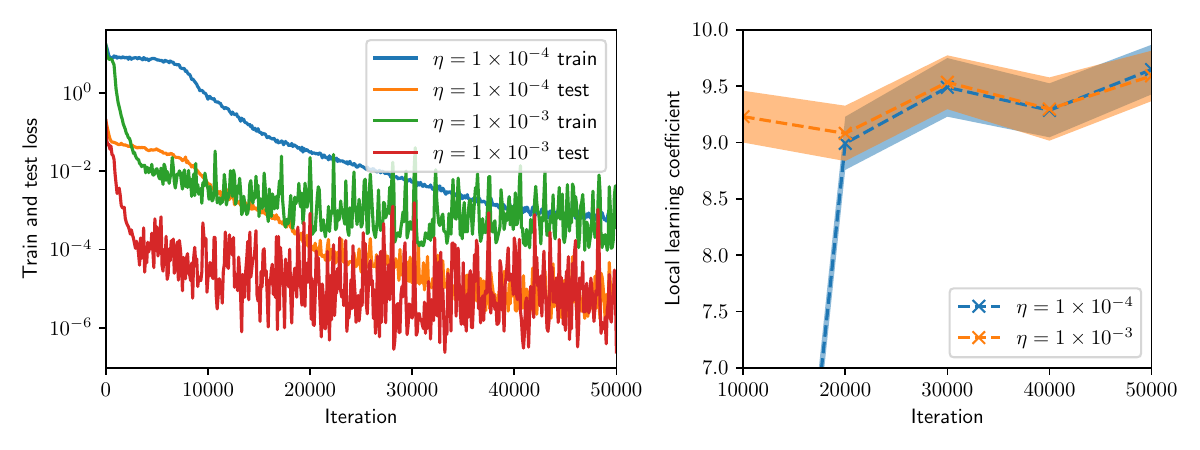}
    \caption{Batch size \(= 8\)}
    \label{fig:bs8}
  \end{subfigure}
  \begin{subfigure}[b]{\textwidth}
    \includegraphics[width=\textwidth, trim={0 0.4cm 0 0}, clip]{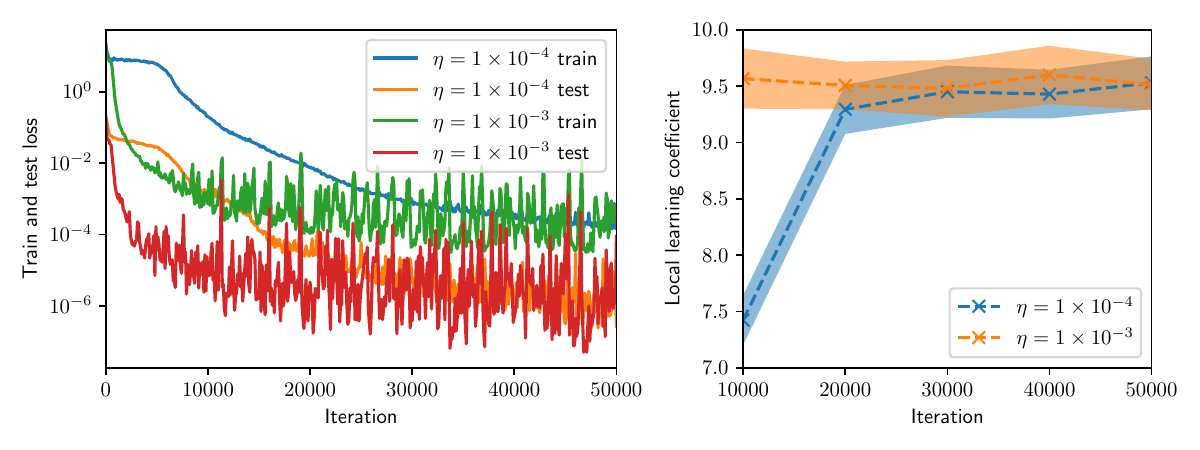}
    \caption{Batch size \(= 16\)}
    \label{fig:bs16}
  \end{subfigure}
  \begin{subfigure}[b]{\textwidth}
    \includegraphics[width=\textwidth, trim={0 0.4cm 0 0}, clip]{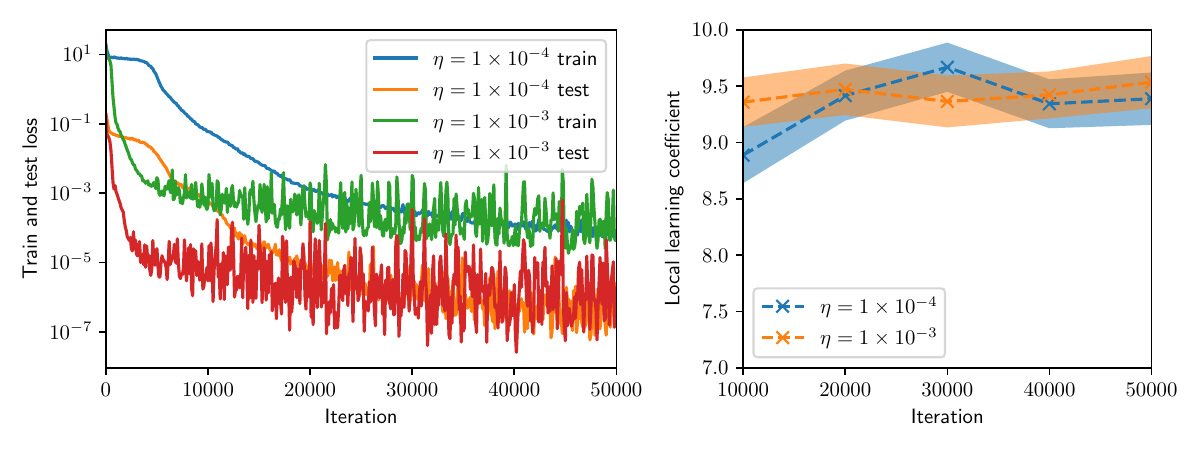}
    \caption{Batch size \(= 32\)}
    \label{fig:bs32}
  \end{subfigure}
  \caption{Training and test histories (left, shown every 100 iterations), and LLC estimates with 95\% confidence intervals (right, shown every 1,000 iterations) for various values of batch size and Adam learning rate.}
  \label{fig:bs}
\end{figure}

\begin{figure}[th]
  \centering
  \begin{subfigure}[b]{\textwidth}
    \includegraphics[width=\textwidth, trim={0 0.4cm 0 0}, clip]{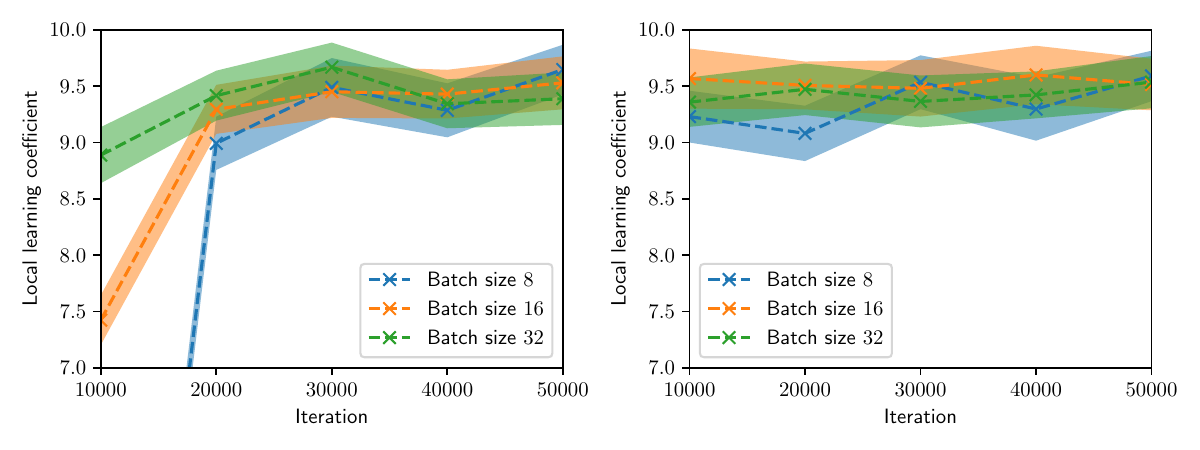}
    \caption{First initialization}
  \end{subfigure}
  \begin{subfigure}[b]{\textwidth}
    \includegraphics[width=\textwidth, trim={0 0.4cm 0 0}, clip]{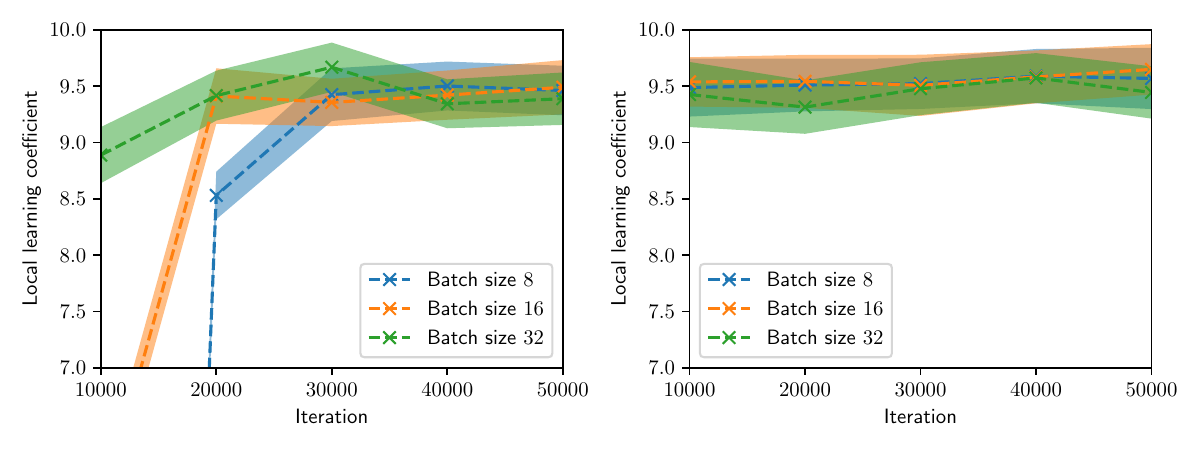}
    \caption{Second initialization}
  \end{subfigure}
  \caption{LLC estimates with 95\% confidence intervals shown every 1,000 iterations, computed for various values of batch size and Adam learning rate \(\eta\) and for two intializations. Left: \(\eta = 1 \times 10^{-4} \). Right: \(\eta = 1 \times 10^{-3} \).}
  \label{fig:lr}
\end{figure}

\Cref{fig:bs} summarizes the results of the computational experiments.
The left panels show the history of the training and test losses every \(100\) iterations.
Here, the test loss is the average of \( (u(x, t, w) - u(x, t))^2\) evaluated over a \( 51 \times 51 \) uniform grid of \((x, t)\) pairs.
The right panels show the LLC estimates computed every \(10,000\) iterations.
Here, care must be taken when interpreting the smaller LLC estimates seen for earlier iterations.
For early iterations, the parameter estimate \(w_t\) is not a minima; therefore, the MCMC chains cover regions of parameter space for which \(L_n(w) < L_n(w_t)\), leading to small and possibly negative values.
Modulo this observation, it can be seen that across all experiments (batch size and learning rate values), the estimated LLC is \(\hat{\lambda}(w^\star) \approx 9.5\).
Additional experiments conducted by varying the pseudo-random number generator seed used to initialize \(w\) also lead to the same estimate of the LLC (see~\cref{fig:lr}.)

This number is remarkable for two reasons: First, despite the estimate of the local minima \(w^\star\) being vastly different across experiments, the LLC estimate is very consistent.
Second, \(\hat{\lambda}(w^\star)\) is significantly smaller than the number of parameters \(d = 20,601\) of the PINN model.
If the PINN model were regular, then it would have a much smaller \(O(10)\) number of parameters.
These results indicate that the PINN solution corresponds to a very flat region of parameter space, and different choices of initializations and stochastic optimization hyperparameters lead to the same region.
As seen in \cref{fig:bs,fig:lr}, the lower learning rate of \(1 \times 10^{-4}\), due to reduced stochastic optimization noise, takes a larger number of iterations to arrive at this region than the larger learning rate of \( 1 \times 10^{-3}.\)

\section{Discussion}
\label{sec:discussion}

A significant limitation of the proposed reformulation of the physics-informed learning problem is that it's limited to PINN models with hard initial and boundary condition constraints.
In practice, physics-informed learning is often carried without such hard constraints by penalizing not only the residual of the governing equation, \( y_\text{GE}(x, t) \coloneqq \mathcal{L}(u(x, t, w), x, t) \) but also the residual of the initial conditions, \(y_{\text{IC}}(x, w) \coloneqq u(x, 0, w)\), and boundary conditions, \(y_{\text{BC}} \coloneqq \mathcal{B}(u(x, t, w), x, t)\).
These residuals can also be regarded as observables of the PINN model.
As such, a possible approach to reformulate the physics-informed learning problem with multiple residuals is to introduce a single vector observable \( y \coloneqq [y_{\text{GE}}, y_{\text{IC}}, y_{\text{BC}}]\) and consider the statistical learning problem problem with model \( p(y \mid x, t, w) q(x, t).\)
In such a formulation, the number of samples \(n\) would be the same for all residuals, which goes against the common practice of using different numbers of residual evaluation points for each term of the PINN loss.
Nevertheless, we note that as indicated in \cref{sec:slt}, PINN learning with randomized batches approximates minimizing the population loss \(L(w) \equiv \lim_{n \to \infty} L_n(w)\), which establishes a possible equivalency between the physics-informed learning problem with multiple residuals and the reformulation described above.
Furthermore, special care must be taken when considering the weighting coefficients commonly used in physics-informed learning to weight the different residuals in the PINN loss.
These problems will be considered in future work.

The presented statistical learning analysis of the physics-informed learning problem has interesting implications on the quantification of uncertainty in PINN predictions.
Bayesian uncertainty quantification (UQ) in PINN predictions is often carried out by choosing a prior on the model parameters and a fixed set of residual evaluation points and computing samples from the posterior distribution of PINN parameters (see, e.g. \cite{zong-2025-randomized,yang-2021-bpinn}.)
As noted in \cite{zong-2025-randomized}, MCMC sampling of the PINN loss is highly challenging computationally due to the singular nature of PINN models, but such an exhaustive exploration of parameter space may not be necessary because the solution candidates may differ significantly in parameter space but correspond to only a limited range of loss values and thus a limited range of possible predictions.
In other words, this analysis justifies taking a function-space view of Bayesian UQ for physics-informed learning due to the singular nature of the PINN models involved.

\begin{figure}[th]
  \centering
  \begin{subfigure}[b]{0.48\textwidth}
    \includegraphics[width=\textwidth, trim={0 0.4cm 0 0}, clip]{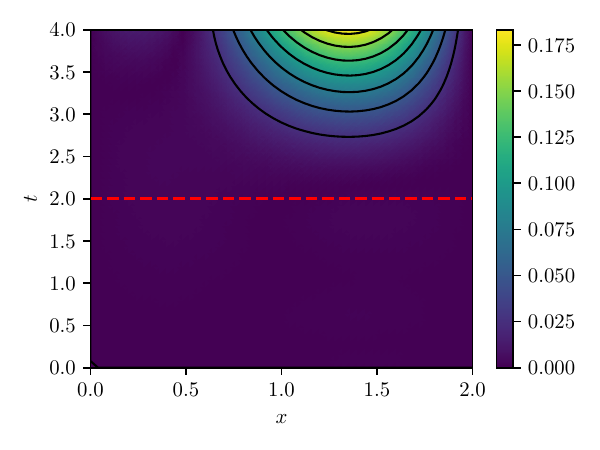}
    \caption{\( | u(x, t) - u(x, t, w^{0}) | \)}
    \label{fig:pred}
  \end{subfigure}
  \centering
  \begin{subfigure}[b]{0.48\textwidth}
    \includegraphics[width=\textwidth, trim={0 0.4cm 0 0}, clip]{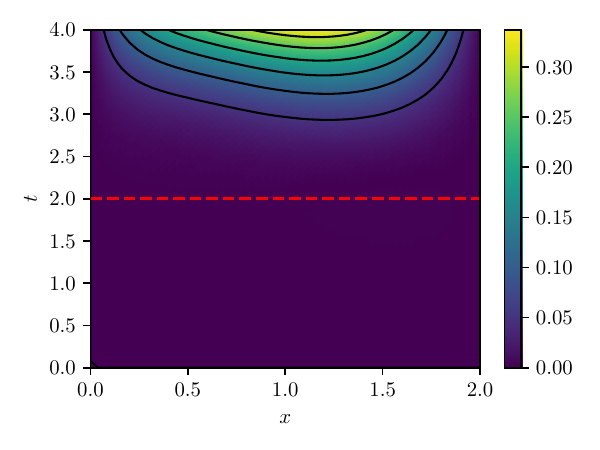}
    \caption{\( | u(x, t) - u(x, t, w^{1}) | \)}
    \label{fig:error}
  \end{subfigure}
  \caption{PINN pointwise absolute estimation errors for two different parameter vectors, \(w^{0}\) and \(w^{1}\), obtained via different initializations of the PINN model parameters. Both parameter estimates were computed using batch size of 32 and Adam learning rate of \(1 \times 10^{-4}\) over \(100,000\) iterations. Despite both cases exhibiting result in similar loss \(L_n(w)\) values (\(O(10^{-5}\)) and similar LLCs \(\hat{\lambda}(w^0) \approx \hat{\lambda}(w^1) \approx 9.5\), they exhibit noticeably different extrapolation behavior.}
  \label{fig:extrapolation_errors}
\end{figure}

Finally, another implication of the presented analysis is related to the extrapolatory capacity of PINNs.
It is understood that PINNs, particularly time-dependent PINNs, exhibit limited extrapolation capacity for times outside the training window \((0, T]\) (see, e.g., \cite{wang-2025-extrapolation}.)
The presented analysis indicates that this is not a feature of the IBVP being solved, or the number and arrangement of residual evaluation points in space and time, or the optimization algorithm, but rather because physics-informed learning estimates model parameters by effectively minimizing \(L(w) \propto \mathbb{E}_{q(x, t)} [\mathcal{L}(u(x, t, w), x, t)]^2\), that is, the expected loss over the input-generating distribution \(q(x, t)\) with support given by the spatio-temporal training windows.
While PINN solution candidates along the flat minima are all within a limited range of loss values, there is no guarantee that this would hold for a different loss \(L'(w) \propto \mathbb{E}_{q'(x, t)} [\mathcal{L}(u(x, t, w), x, t)]^2\) because the structure of the flat minima is tightly tied to the definition of the original loss, which is tightly tied to the input-generating distribution and thus to the spatio-temporal training window.
In fact, it can be expected that two \(w\)s with losses \(L(w)\) within tolerance will lead to vastly different extrapolation losses \(L'(w)\) (e.g., see \cref{fig:extrapolation_errors}), consistent with extrapolation results reported in the literature.

\section{Conclusions}
\label{sec:conclusions}

We have presented in this work a reformulation of physics-informed learning for IBVPs as a statistical learning problem valid for hard-constraints parameterizations.
Employing this reformulation we argue that the physics penalty in physics-informed learning is not a regularizer as it is often interpreted (see, e.g. \cite{karniadakis-2021-physics}) but as an infinite source of data, and that physics-informed learning with randomized batches of residual evaluation points is equivalent to minimizing \(L(w) \propto \mathbb{E}_{q(x, t)} [\mathcal{L}(u(x, t, w), x, t)]^2\).
Furthermore, we show that the PINN loss landscape is not characterized by discrete, measure-zero local minima even in the limit \(n \to \infty\) of the number of residual evaluation points but by very flat local minima corresponding to a low-complexity (relative to the number of model parameters) model class.
We characterize the flatness of local minima of the PINN loss landscape by computing the LLC for a heat equation IBVP and various initializations and choices of learning hyperparameters, and find that across all initializations and hyperparameters studied the estimated LLC is very similar, indicating that the neighborhoods around the different PINN parameter estimates obtained across the experiments exhibit similar flatness properties.
Finally, we discuss the implications of the proposed analysis on the quantification of the uncertainty in PINN predictions, and the extrapolation capacity of PINN models.

\section*{Acknowledgments}
This material is based upon work supported by the U.S. Department of Energy (DOE), Office of Science, Office of Advanced Scientific Computing Research. Pacific Northwest National Laboratory is operated by Battelle for the DOE under Contract DE-AC05-76RL01830.

\bibliographystyle{elsarticle-num}
\bibliography{refs}

\end{document}